%% file: main.tex
\documentclass[10pt,twocolumn,letterpaper]{article}

\usepackage{cvpr}              % To produce the CAMERA-READY version
\input{preamble}
\definecolor{cvprblue}{rgb}{0.21,0.49,0.74}
\usepackage[pagebackref,breaklinks,colorlinks,allcolors=cvprblue]{hyperref}

\def\paperID{4736} % *** Enter the Paper ID here
\def\confName{CVPR}
\def\confYear{2026}

\title{Learning to Restore Multi-Degraded Images via Ingredient Decoupling and Task-Aware Path Adaptation}

\author{Hu Gao\\
Shanghai Jiao Tong University  \\
{\tt\small gao\_h@sjtu.edu.cn}
\and
Xiaoning Lei\\
CATL\\
{\tt\small leixn01@outlook.com}
\and
Ying Zhang\\
Beijing Normal University\\
{\tt\small 202331081001@mail.bnu.edu.cn}
\and
Xichen Xu\\
Shanghai Jiao Tong University  \\
{\tt\small neptune\_2333@sjtu.edu.cn}
\and
Guannan Jiang\\
CATL\\
{\tt\small jianggn@catl.com}
\and
Lizhuang Ma$^*$\\
Shanghai Jiao Tong University  \\
{\tt\small lzma@sjtu.edu.cn}
}

\begin{document}
\maketitle
\input{sec/0_abstract}    
\input{sec/1_intro}

\input{sec/2_relate}

\input{sec/3_method}

\input{sec/4_exper}
\input{sec/5_con}

{
    \small
    \bibliographystyle{ieeenat_fullname}
    \bibliography{main}
}

\input{sec/X_suppl}

\end{document}

%% file: sec/0_abstract.tex
\begin{abstract}
Image restoration (IR) aims to recover clean images from degraded observations. Despite remarkable progress, most existing methods focus on a single degradation type, whereas real-world images often suffer from multiple coexisting degradations, such as rain, noise, and haze coexisting in a single image, which limits their practical effectiveness. 
In this paper, we propose an adaptive multi-degradation image restoration network that reconstructs images by leveraging decoupled representations of degradation ingredients to guide path selection. Specifically,  we design a degradation ingredient decoupling block (DIDBlock) in the encoder to separate degradation ingredients statistically by integrating spatial and frequency domain information, enhancing the recognition of multiple degradation types and making their feature representations independent. In addition, we present fusion block (FBlock) to integrate degradation information across all levels using learnable matrices. In the decoder, we further introduce a task adaptation block (TABlock) that dynamically activates or fuses functional branches based on the multi-degradation representation, flexibly selecting optimal restoration paths under diverse degradation conditions. The resulting tightly integrated architecture, termed IMDNet, is extensively validated through experiments, showing superior performance on multi-degradation restoration while maintaining strong competitiveness on single-degradation tasks.
\end{abstract}

%% file: sec/1_intro.tex
\section{Introduction}
\label{sec:intro}
Image restoration (IR) is the process of reconstructing high-quality images from degraded observations. It is a typical ill-posed problem, where multiple potential solutions exist for the same degraded input. Traditional methods~\cite{2011Single, 10558778} address this by introducing task-specific priors, formulating degradation models, and applying inverse operations. Although effective in certain cases, these approaches rely on strong assumptions about degradation factors such as noise or blur kernels. In real-world scenarios, however, degradation processes are uncertain and often unknown, making accurate modeling difficult and leading to poor generalization and unstable performance.

In recent years, deep neural networks~\cite{PGH2Netisu2025prior,FSNet, VLUNetZeng_2025_CVPR} have achieved remarkable progress in image restoration. By capturing the statistical properties of natural images, deep learning-based methods can implicitly learn a wide range of priors, leading to superior performance over traditional approaches. Based on the number of degradation types a single model can handle, image restoration methods are typically categorized into task-specific, task-aligned, and all-in-one approaches. Task-specific methods~\cite{PGH2Netisu2025prior, efderainguo2025efficientderain+} design specialized models for individual tasks, often tailored to the characteristics of the target data. Task-aligned methods~\cite{FSNet, aclgu2025acl} aim to build a general network trained sequentially on datasets of different degradation types, which can then be applied to multiple restoration tasks. All-in-one methods~\cite{potlapalli2023promptir,VLUNetZeng_2025_CVPR} share the goal of designing a general network but differ in that the model is trained simultaneously on multiple degradation types, enabling a single network to handle diverse degradations.

\begin{figure} % use float package if you want it here
    \centerline{\includegraphics[width=1\linewidth]{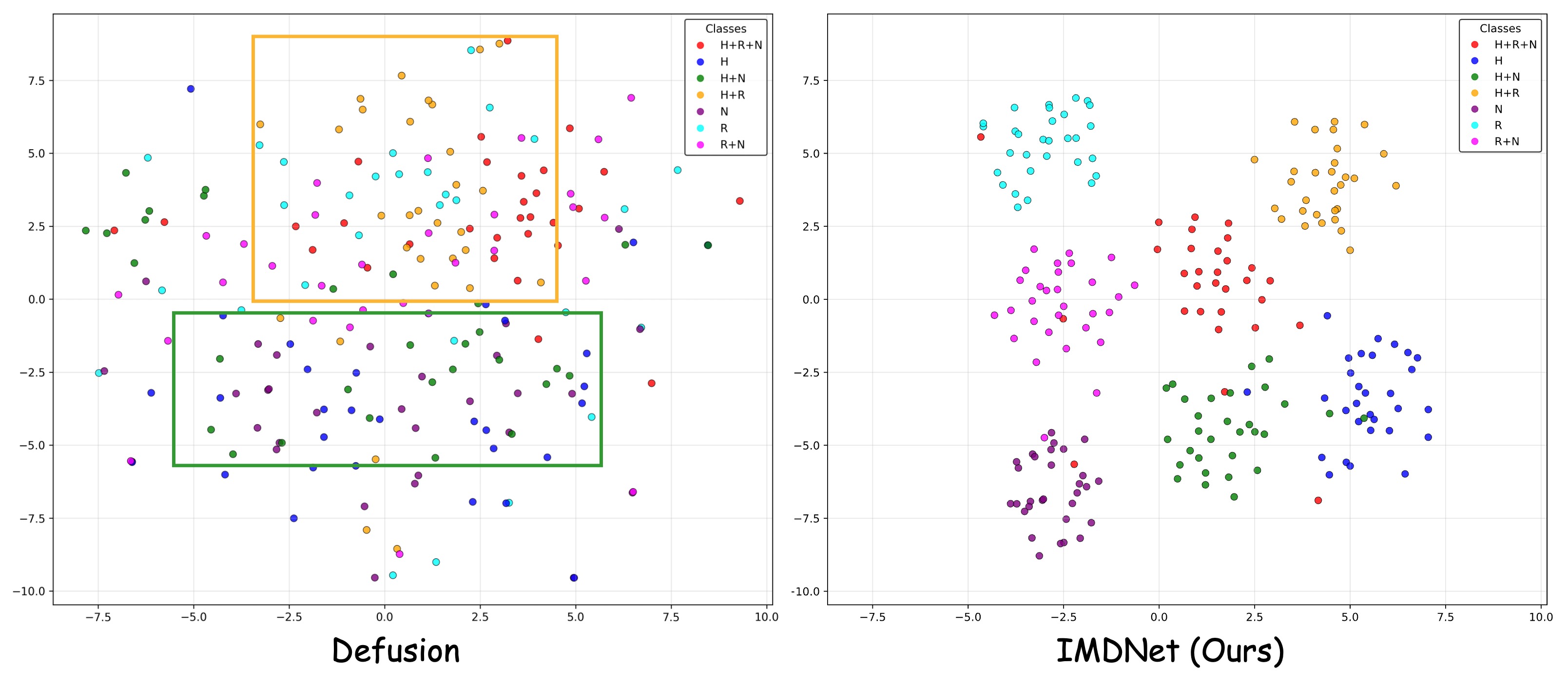}}
	\caption{The figure presents t-SNE visualizations of degradation embeddings from IMDNet (ours) and Defusion~\cite{DefusionLuo_2025_CVPR}. Our model exhibit clearer clustering, highlighting the effectiveness of decomposing degradation ingredients into decoupled representations.}
 \label{fig:tsne}
\end{figure}

However, the above methods~\cite{FSNet, VLUNetZeng_2025_CVPR} focus on single-degradation image restoration (SDIR) and generally assume that each image contains only one type of degradation, such as rain, haze, or noise. In contrast, real-world scenes are highly heterogeneous, dynamic, and uncertain, with image degradations often forming complex combinations, for example, rain, haze, and noise coexisting in a single image. The intricate nature of multi-degradation scenarios hampers the performance of existing methods.

Although some studies~\cite{AASO,FDTANetgao2025frequency,tanet,OWN,DIGNet} have explored multi-degradation image restoration (MDIR), their performance remains limited, partly because they rely solely on simple attention mechanisms to adaptively select useful features. Ref-IRT~\cite{REF-IRT} addresses this limitation with a multi-stage framework that progressively transfers similar edges and textures from reference images. However, this approach ignores the interaction between different degradation mechanisms.
In addition, several All-in-one methods~\cite{DefusionLuo_2025_CVPR, Perceive-IR10990319} have achieved promising results on multi-degraded images by leveraging large vision models. Nevertheless, as shown in Figure~\ref{fig:tsne}, although the retrained Defusion~\cite{DefusionLuo_2025_CVPR} model can roughly distinguish degradation ingredients (highlighted by the yellow and green boxes), the separation is still weak. It struggles to learn intrinsic differences among degradation types, instead relying on its strong representational power to fit mappings from low- to high-quality images.

\begin{figure} % use float package if you want it here
    \centerline{\includegraphics[width=1\linewidth]{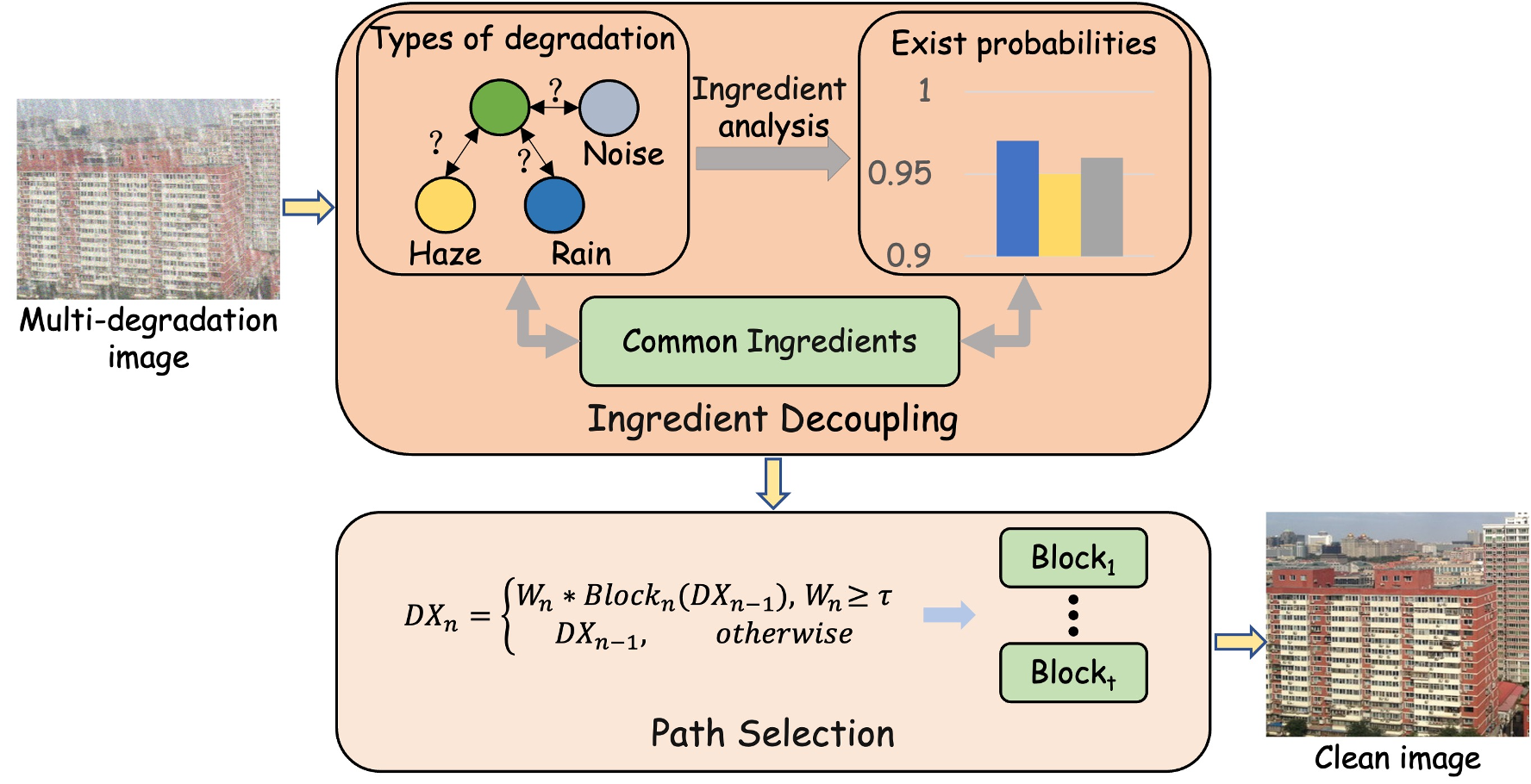}}
	\caption{Mechanisms of our method. Our approach achieves multi-degradation image restoration by decomposing degradation ingredients into decoupled representations.}
 \label{fig:exam}
\end{figure}

Given these considerations, a natural question arises: is it possible to design a network capable of adaptively restoring images affected by multiple degradations according to variations in their constituent ingredients? To address this challenge, we propose IMDNet (Figure~\ref{fig:exam}), an adaptive MDIR network that reconstructs high-quality images by  leveraging decoupled representations of degradation ingredients to guide path selection. In the encoder, we design a degradation ingredient decoupling block (DIDBlock) statistically separates degradation ingredients using spatial and frequency information, enhancing recognition of multiple degradation types while keeping their feature representations independent. To further consolidate information, we incorporate a fusion block (FBlock) that adaptively integrates degradation features across all hierarchical levels using learnable matrices, facilitating comprehensive representation of complex degradations.
In the decoder, we introduce a task adaptation block (TABlock) to dynamically activate or fuse functional branches according to the multi-degradation representation, allowing the network to flexibly select the optimal restoration path under varying degradation conditions. Figure~\ref{fig:tsne} shows that IMDNet leverages the "degradation ingredient $\rightarrow$ path selection" mechanism for accurate identification of different degradation types.

The main contributions of this work are:
\begin{enumerate}
	\item We propose IMDNet for MDIR, leveraging decoupled representations of degradation ingredients to guide path selection. Extensive experiments show its superior performance on both MDIR and SDIR.
    \item We design a degradation ingredient decoupling block (DIDBlock)  to decompose degradation ingredients statistically using spatial and frequency information, improving recognition of multiple degradation types while maintaining independent feature representations.
	\item We introduce a task adaptation block (TABlock) that dynamically activates or fuses branches based on the multi-degradation representation, flexibly selecting optimal restoration paths under varying conditions.
    \item We incorporate a fusion block (FBlock) to combine degradation information from all levels using learnable matrices.
\end{enumerate}

%% file: sec/2_relate.tex
\section{Related Works}
\label{sec:relate}

\subsection{Single-degradation Image Restoration}
SDIR focuses on restoring high-quality images degraded by a single specific type of corruption. Traditional approaches~\cite{2011Single, 10558778} typically address this ill-posed problem by introducing handcrafted priors to constrain the solution space. However, these priors rely heavily on expert experience and often suffer from limited adaptability and poor generalization to diverse degradation scenarios.

With the rapid advancement of deep learning in high-level vision tasks, numerous data-driven methods~\cite{LSSRgao2024learning,adarevD10656920,FSNet,aclgu2025acl,VLUNetZeng_2025_CVPR} have been developed for image restoration. Depending on the number of degradation types a single model can handle, these methods are generally classified into task-specific, task-aligned, and all-in-one approaches.

Task-specific methods~\cite{PGH2Netisu2025prior, efderainguo2025efficientderain+} design dedicated models for each degradation type, often customized to the characteristics of the corresponding dataset.
ALGNet~\cite{ALGgao2024learning} introduces a local feature extraction module to mitigate local pixel forgetting caused by excessive hidden units, achieving efficient and precise deblurring.
XYScanNet~\cite{liu2024xyscannet} adopts an alternating intra- and inter-slice scanning strategy to better capture spatial dependencies.
EfDeRain+~\cite{efderainguo2025efficientderain+} redefines deraining as a predictive filtering task, avoiding the need for complex rain modeling assumptions.
UPID-EDM~\cite{upid10.1145/3664647.3680560} employs a diffusion-based unpaired learning framework, leveraging vision-language priors and energy-guided sampling to retain structural content while effectively removing degradations.
PGH$^2$Net~\cite{PGH2Netisu2025prior} reveals how bright/dark channel priors and histogram equalization jointly guide hierarchical feature learning for dehazing.
Diff-Unmix~\cite{diffunmix10656884} performs self-supervised denoising through spectral decomposition and conditional diffusion modeling.

Task-aligned methods~\cite{FSNet, aclgu2025acl} seek to develop unified networks capable of handling multiple degradation types by sequentially training on different restoration datasets.
MambaIR~\cite{guo2025mambair} introduces a four-directional unfolding strategy combined with channel attention to enhance spatial representation, while MambaIRV2~\cite{guo2025mambairv2} further integrates non-causal modeling inspired by ViTs to strengthen state space expressiveness.
FSNet~\cite{FSNet} designs dynamic frequency selection modules to adaptively extract the most informative spectral components for restoration.
ACL~\cite{aclgu2025acl} leverages the mathematical equivalence between linear attention and SSM in Mamba, unlocking the potential of linear attention for image restoration.
MHNet~\cite{gao2025mixed} adopts a mixed hierarchical design to produce restorations with richer textures and finer structural details.

All-in-one methods~\cite{potlapalli2023promptir,VLUNetZeng_2025_CVPR} aim to build a unified network capable of handling multiple degradation types simultaneously, allowing a single model to manage diverse corruption scenarios.
PromptIR~\cite{potlapalli2023promptir} proposes a prompt-driven restoration framework that recovers clean images directly from inputs using lightweight, plug-and-play prompt modules.
NDR~\cite{NDR10680296} learns neural degradation representations to capture shared characteristics across different degradation types.
VLU-Net~\cite{VLUNetZeng_2025_CVPR} leverages vision-language model features to automatically identify degradation-aware keys, removing the need for manually defined degradation categories.
AutoDIR~\cite{autodir10.1007/978-3-031-73661-2_19} combines VLM guidance with latent diffusion to detect unknown degradations and perform structure-consistent restoration.
AdaIR~\cite{cui2025adair} separates degradation from clean image content by jointly exploiting spatial and frequency domain.

Despite their effectiveness in SDIR, these methods encounter difficulties in complex MDIR, where intertwined degradations severely impair restoration accuracy.
Although recent all-in-one approaches~\cite{DefusionLuo_2025_CVPR, Perceive-IR10990319} have leveraged large vision models to enhance performance on multi-degraded images, their ability to distinguish degradation components remains limited. Instead of learning the intrinsic differences among various degradation types, these models primarily depend on strong representational power to directly map low-quality inputs to high-quality outputs.

\subsection{Multi-degradation Image Restoration}
MDIR aims to restore a single image affected by multiple degradation types into a clean image. AASO~\cite{AASO} performs multiple restoration operations in parallel and uses attention to weight each operation, selecting the most suitable restoration strategy. However, the heterogeneity of features generated by different operations limits its performance on multi-degradation images. OWAN~\cite{OWN} employs high-order tensor fusion to coordinate heterogeneous features and leverage high-order statistical information, but its complex network struggles to restore high-frequency details. DIGNet~\cite{DIGNet} considers sequential and spatially varying distortions, while MEPSNet~\cite{Kim_2020_ACCV} uses a mixture-of-experts with parameter sharing to address varying distortions across image regions.

Despite these efforts, performance remains limited, partly because they depend solely on simple attention mechanisms to adaptively select features. To overcome this, Ref-IRT~\cite{REF-IRT} introduces a three-stage restoration framework: the first stage estimates residuals in a coarse-to-fine manner, while the second and third stages progressively transfer details from reference images. FDTANet~\cite{FDTANetgao2025frequency} performs adaptive restoration in the frequency domain based on component differences. However, both methods overlook the interactions between different degradation mechanisms. In this paper, we first separate degradation ingredients by integrating spatial and frequency information, improving recognition of multiple degradation types and making their features more independent. Then, based on multi-degradation representations, we dynamically activate or fuse branches to flexibly select optimal restoration paths, achieving effective multi-degradation image restoration.

%% file: sec/3_method.tex
\section{Method}
\label{sec:method}
In this section, we first outline the overall IMDNet framework, then detail the proposed degradation ingredient decoupling block (DIDBlock) and task adaptation block (TABlock), followed by the loss function.

\begin{figure*} % use float package if you want it here
    \centerline{\includegraphics[width=1\linewidth]{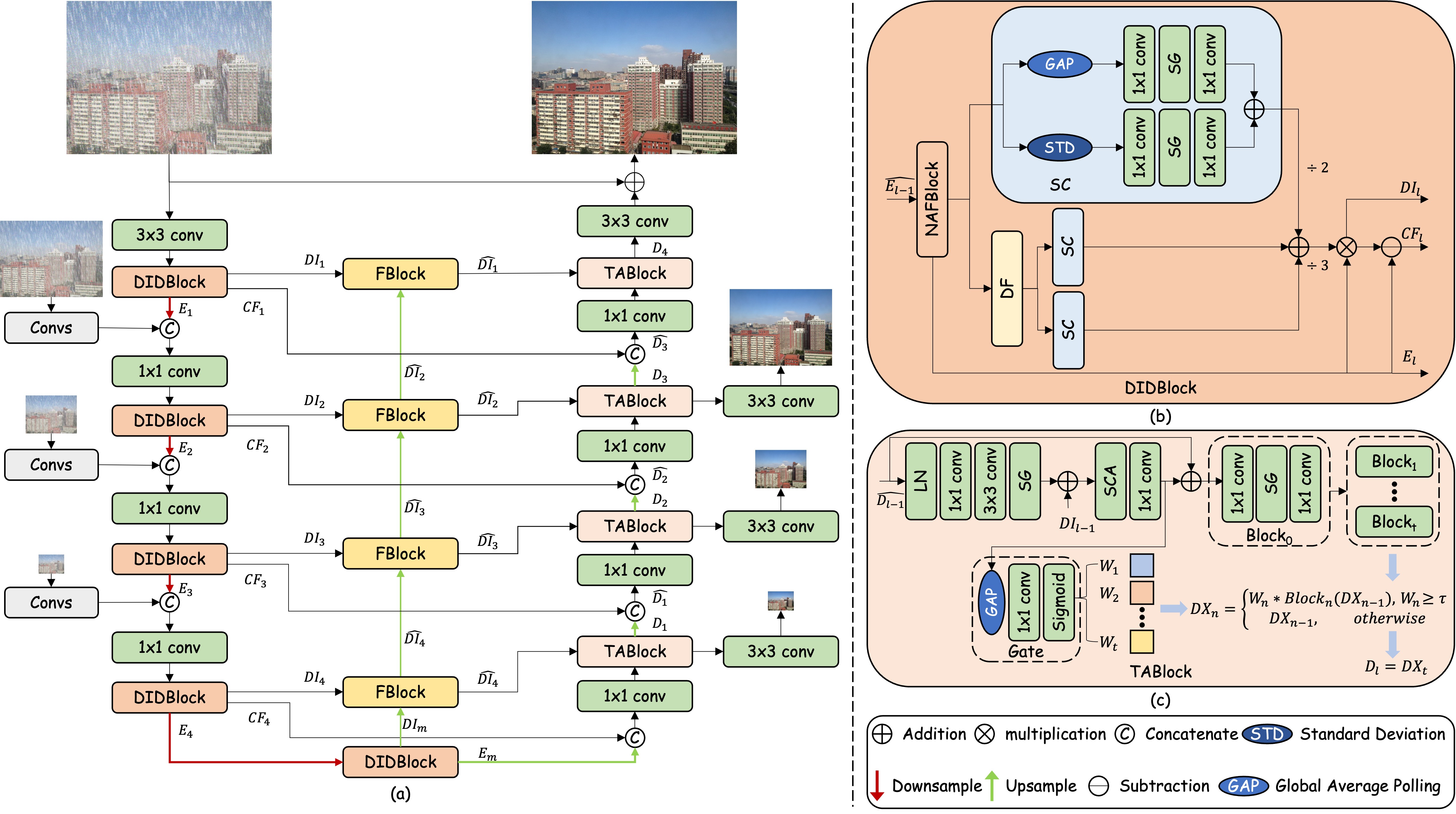}}
	\caption{(a) The overall architecture of the proposed IMDNet. (b) The structure of degradation ingredient decoupling block (DIDBlock). (c) The structure of task adaptation block (TABlock).}
 \label{fig:network}
\end{figure*}

\subsection{Overall Pipeline} 
Our proposed IMDNet, illustrated in Figure~\ref{fig:network}, adopts an encoder–decoder architecture. 
Each encoder level comprises [4, 4, 4, 8] DIDBlocks, while the middle block includes 8 DIDBlocks. Each decoder level consists of [2, 2, 2, 2] TABlocks. Given a degraded image $\mathbf{I} \in \mathbb{R}^{H \times W \times 3}$, IMDNet first applies convolution to extract shallow features $\mathbf{F_{0}} \in \mathbb{R}^{H \times W \times C}$ ($H$, $W$, and $C$ denote the height, width, and number of channels, respectively). These shallow features are then processed through a four-level encoder and a middle block. Each encoder level and the middle block output three types of features: the encoded feature $E$ for the next-level encoder, the decoupled degradation information $DI$, and the skip feature $CF$ used to facilitate the reconstruction.
To enhance training, IMDNet adopts multi-input and multi-output mechanisms. Low-resolution degraded images are introduced into the main path through a Convs consisting of multiple convolutions and concatenations, followed by a channel-adjustment convolution.
The deepest features are subsequently passed into a four-scale decoder that progressively restores spatial resolution. Throughout this process, the degradation information $DI$ is integrated across all levels using learnable matrices within the fusion block (FBlock). Finally, refined features are used to generate a residual image $\mathbf{I_R} \in \mathbb{R}^{H \times W \times 3}$, which is added to the degraded input to produce the restored image: $\mathbf{\hat{I}} = \mathbf{I_R} + \mathbf{I}$.

\subsection{Degradation Ingredient Decoupling Block} 

The features of degraded images entering the encoder naturally contain degradation information, which motivates us to explore whether we can decouple degraded features from clean image features during encoding. The clean features are then used for skip connections between the encoder and decoder, while the degraded features estimate the probabilities of various degradation factors and guide the decoder to perform adaptive multi-degradation image restoration.

To verify this idea, as shown in Figure~\ref{fig:network}(b), we design a degradation ingredient decoupling block (DIDBlock). First, we use the NAFBlock~\cite{chen2022simple} to extract spatial-domain features. Considering the distinct characteristics of different degradation components across spatial and frequency domains, we use a learnable dynamic filter (DF) to separate frequency components and obtain high- and low-frequency features. Finally, we adopt a joint frequency–spatial analysis strategy based on statistical coefficients (SC) to achieve degradation ingredient decoupling representation, effectively isolating and characterizing degradation information from both spatial and frequency features. Specifically, for the feature $E_{l-1}$ output by the encoder at level $l-1$, we first concatenate it with the low-resolution degraded image features and adjust its dimension using a $1 \times 1$ convolution to obtain $\widehat{E_{l-1}}$. We then feed $\widehat{E_{l-1}}$ into the NAFBlock to generate the spatial-domain feature $E_l$ for the level-$l$ encoder:
\begin{equation}
\label{eq:naf}
E_l = NAFBlock(f_{1 \times 1}^c([E_{l-1},Convs(I_l)]))
\end{equation}
where $f_{1 \times 1}^c$ represents the $1 \times 1$ convolution, and $[\cdot]$ is the concatenation operation. Next, we apply the dynamic filter to decompose $E_l$ into high-frequency and low-frequency components:
\begin{equation}
\label{eq:df}        
F_{H}, F_{L} = DF(E_l) 
\end{equation}

We then perform statistical analysis of the spatial features $E_l$ and frequency domain features $F_{H}$, $F_{L}$ to isolate the degradation information:
\begin{equation} 
\begin{aligned}
\label{eq:sc}   
DI_l &= \frac{SC(F_{H}) \oplus SC(F_{L})  \oplus SC(E_l)}{6} \otimes E_l \\
\\
SC(\cdot) &= f_{1 \times 1}^c(SG(f_{1 \times 1}^c(GAP(\cdot)))) 
\\ & \oplus f_{1 \times 1}^c(SG(f_{1 \times 1}^c(STD(\cdot))))
\end{aligned}
\end{equation}
where $GAP(\cdot)$ and $STD(\cdot)$ represent global average pooling and standard deviation pooling, respectively. $SG(\cdot)$ indicates the simple gate, serving as a replacement for the nonlinear activation function. Given an input $X_{f} \in \ R^{H \times W \times C}$, SG splits it into two features $ X_{f}^{1}, X_{f}^{2} \in \ R^{H \times W \times \frac{C}{2}}$ and calculates $X_{f}^{1}, X_{f}^{2}$ using a linear gate., Finally, we obtain the clean features $CF_l$ by subtracting the degradation information from the encoded features:
\begin{equation}
\label{eq:cf}
CF_l = E_l \ominus DI_l
\end{equation}

Through this process, we obtain three key outputs: the feature $E_l$ for the next-level encoder, the feature $CF_l$ for skip connections, and the degradation information $DI_l$ to guide adaptive multi-degradation restoration in the decoder.
This design offers several advantages. By leveraging the statistical properties of input features, the generated degradation information captures both the image’s content correlation and shared ingredient characteristics, improving representation consistency and knowledge transfer across different degradation types. Moreover, unlike previous methods that use simple addition or concatenation for skip connections, our approach effectively suppresses degraded information from the encoder and minimizes interference from implicit noise.

\subsection{Task Adaptation Block} 

To enable the model to dynamically activate or fuse multiple functional branches based on the decoupled degradation information, flexibly select the optimal restoration path, and adaptively handle various combinations of degraded images, we design a task adaptive block (TABlock) in the decoder. As shown in Figure~\ref{fig:network}(c), TABlock first captures contextual information features and then fuses the degradation information. Notable, the first branch serves as a general branch, responsible for processing common components shared across all degradation types and ensuring stable baseline restoration performance. To achieve dynamic branch fusion guided by task relevance, we employ a gated network that adaptively generates activation weight vectors for each branch. To further enhance computational efficiency, we introduce a branch sparsity constraint to achieve sparse activation, retaining only branches with significant weights for computation. Finally, the outputs of all active branches are weighted according to their activation weights to produce the final restored features. 

Specifically, for the feature $\widehat{D_{l-1}}$, obtained by fusing the decoder output and skip-connection features at level $l-1$, we first extract contextual information features and then inject the degradation information $\widehat{DI_{l-1}}$ to obtain the feature $DC_{l-1}$:
\begin{equation}
\label{eq:dc}
DC_{l-1} = f_{1 \times 1}^c(SCA(SG(f_{3 \times 3}^{dwc}(f_{1 \times 1}^c(LN(\widehat{D_{l-1}}))) \oplus \widehat{DI_{l-1}})))
\end{equation}
where $f_{3 \times 3}^{dwc}$ denotes the $3 \times 3$ depth-wise convolution, $SCA(\cdot)$ is the simplified channel attention~\cite{chen2022simple}, and $LN(\cdot)$ represents layer normalization.
Notably, $\widehat{DI_{l-1}}$ is obtained by aggregating degradation information from all levels through the fusion block (FBlock) as:
\begin{equation}
\label{eq:fblock}
\widehat{DI_{l-1}} = DI_l \oplus W\widehat{DI_{l}}
\end{equation}
where $W$ denotes learnable parameters that are directly optimized via backpropagation and initialized to 1. Next, for the first general branch, the computation is defined as:
\begin{equation}
\begin{aligned}
\label{eq:gb}
DX_0 &= Block_0(DC_{l-1} \oplus \widehat{D_{l-1}})
\\
Block(\cdot) &= f_{1 \times 1}^c(SG(f_{1 \times 1}^c(\cdot))).
\end{aligned}
\end{equation}

Subsequently, we feed $DC_{l-1}$ into a gated network to generate the activation weight vector for each branch:
\begin{equation}
\label{eq:gw}
W_{n} = Sigmoid(f_{1 \times 1}^c(GAP(DC_{l-1}))).
\end{equation}

To further enhance efficiency, we apply a branch sparsity constraint to retain only branches with significant weights for computation:
\begin{equation} 
\label{eq:bs} 
DX_n=\left\{ 
    \begin{aligned} 
        &W_n \ast Block_n(DX_{n-1}), W_n \geq \tau, \\ &DX_{n-1} , otherwise. 
    \end{aligned} 
    \right. 
\end{equation}
where $Block_n(\cdot)$ represents the $n$-th functional branch, and $\tau$ is a threshold empirically set to 0.2.

\begin{table*}
\centering
\caption{Quantitative results of different models conducted on various combinations of degradation types, where H, R, and N denote haze, rain, and noise, respectively.}
\label{tb:mix}
\resizebox{\linewidth}{!}{
\begin{tabular}{cccccccc||c}
    \hline
   Methods& H + R + N  & H + R &H + N& R + N & H &R&N&Average
    \\
    \hline\hline
    Restormer~\cite{Zamir2021Restormer} &23.52/0.792 &25.45/0.836 &26.73/0.863	&25.31/0.815	&29.82/0.931	&28.05/0.873	&27.44/0.837 &26.62/0.850
     \\
    AirNet~\cite{all_conli2022all} &27.41/0.812 &28.28/0.889 &27.59/0.861 &26.98/0.821 &30.22/0.959 &28.27/0.881 & 27.25/0.847 &28.00/0.867
       \\
    U$^2$Former~\cite{u2former} &25.07/0.803 &26.23/0.856 &26.79/0.864	&26.02/0.816	&29.95/0.933	&28.50/0.876	&27.12/0.831 &27.09/0.854
     \\
    PromptIR~\cite{potlapalli2023promptir} &27.54/0.819	&28.43/0.901 &27.99/0.871	&27.05/0.822	&30.46/0.956	&28.78/0.885	&27.92/0.851 &28.31/0.872
       \\
    VLU-Net~\cite{VLUNetZeng_2025_CVPR} &29.35/0.842	&30.38/0.935 &28.79/0.869	&28.21/0.843	&31.37/0.959	&33.82/0.968	&27.88/0.853 &29.97/0.896
       \\
FDTANet~\cite{FDTANetgao2025frequency}  &29.68/0.846	&30.91/0.937	&\textbf{29.06}/\underline{0.874}	&27.63/0.827	&\textbf{31.91/0.981} &29.13/0.892	&28.88/0.857 &29.60/0.888
\\
Ref-IRT~\cite{REF-IRT}&28.95/0.823	&\underline{32.33/0.961} &28.62/0.872	&31.22/0.831	&30.69/0.961	&31.89/0.902	&29.39/0.899 &30.44/0.893
\\
Perceive-IR~\cite{Perceive-IR10990319}&\underline{30.22/0.894}	&31.55/0.945	&28.32/0.871	&30.52/0.849	&\underline{31.90/0.980}	&38.95/0.983	&31.88/0.923	&31.90/0.921
\\
Defusion~\cite{DefusionLuo_2025_CVPR}&29.66/0.849	&29.93/0.937	&28.11/0.866	&\textbf{32.17/0.920}	&31.29/0.969	&\textbf{40.15}/\underline{0.986}	&\underline{32.72/0.925}	&\underline{32.01/0.922}
\\
\hline
\textbf{IMDNet(Ours)} &\textbf{31.19/0.910}	&\textbf{33.21/0.973}	&29.01/\textbf{0.895}	&\underline{32.15/0.918}	&31.73/0.972	&\underline{40.14}/\textbf{0.988}	&\textbf{33.86/0.932}	&\textbf{33.04/0.941}

    \\
    \hline
\end{tabular}}
\end{table*}

\begin{figure*} % use float package if you want it here
    \centerline{\includegraphics[width=1\linewidth]{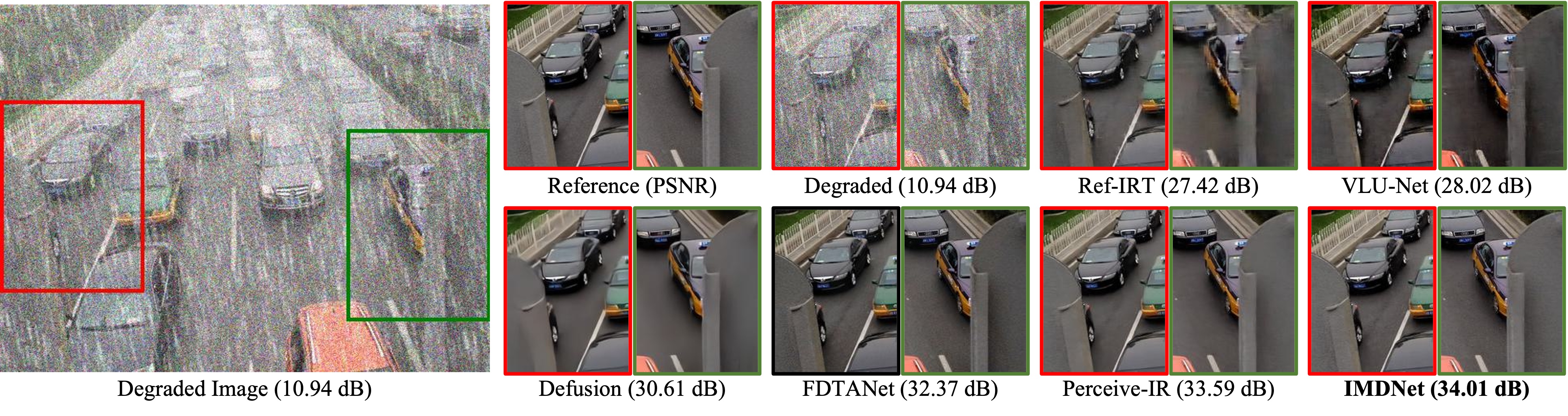}}
	\caption{Qualitative results under the MDIR experimental setup. Compared to other methods, our IMDNet effectively reduces color distortion and produces images that are visually closer to the ground truth.}
 \label{fig:mix}
\end{figure*}

\subsection{Loss Function}
Consistent with prior works, we optimize IMDNet in both spatial and frequency domains. To facilitate accurate decoupling in the encoder, the skip-connection feature $CF$ is encouraged to contain minimal degradation information, while the branch-selection feature $DI$ is designed to capture primarily degradation information. We enforce this separation using a cosine similarity loss, maximizing the distance between 
$CF$ and $DI$. The overall loss function is defined as:
\begin{equation}
\begin{aligned}
\label{eq:loss1}
L &= \sum_{i=1}^{4}(L_{c}(\hat{I_i},\overline I_i)  + \delta L_{e}(\hat{I_i},\overline I_i) 
\\
& + \lambda L_{f}(\hat{I_i},\overline I_i) + \gamma L_{d}(CF_i, DI_i))
\\
L_{c} &= \sqrt{||\hat{I_i} -\overline I_i||^2 + \epsilon^2}
\\
L_{e} &= \sqrt{||\triangle \hat{I_i} - \triangle \overline I_i||^2 + \epsilon^2}
\\
L_{f} &= ||\mathcal{F}(\hat{I}_i)-\mathcal{F}(\overline I_i)||_1
\\
L_{d} & = \frac{CF_i \cdot DI_i}{\|CF_i\|_2 \, \|DI_i\|_2}
\end{aligned}
\end{equation}
where $i$ indexes input/output images at different scales, $\overline I_i$ denotes the target images and $L_{c}$ is the  Charbonnier loss with constant $\epsilon$ empirically set to $0.001$ for all the experiments. $L_{e}$ is the edge loss, where $\triangle$ represents the  Laplacian operator. $L_{f}$  denotes the frequency domains loss, where $\mathcal{F}$ represents fast Fourier transform. $L_{d}$ is the decoupling loss.  The weights $\lambda$,  $\delta$ and $\gamma$ balance the loss terms and are set to 0.1, 0.05, and 0.001, respectively, following~\cite{Zamir2021MPRNet,FSNet}.

%% file: sec/4_exper.tex
\begin{figure*} % use float package if you want it here
    \centerline{\includegraphics[width=1\linewidth]{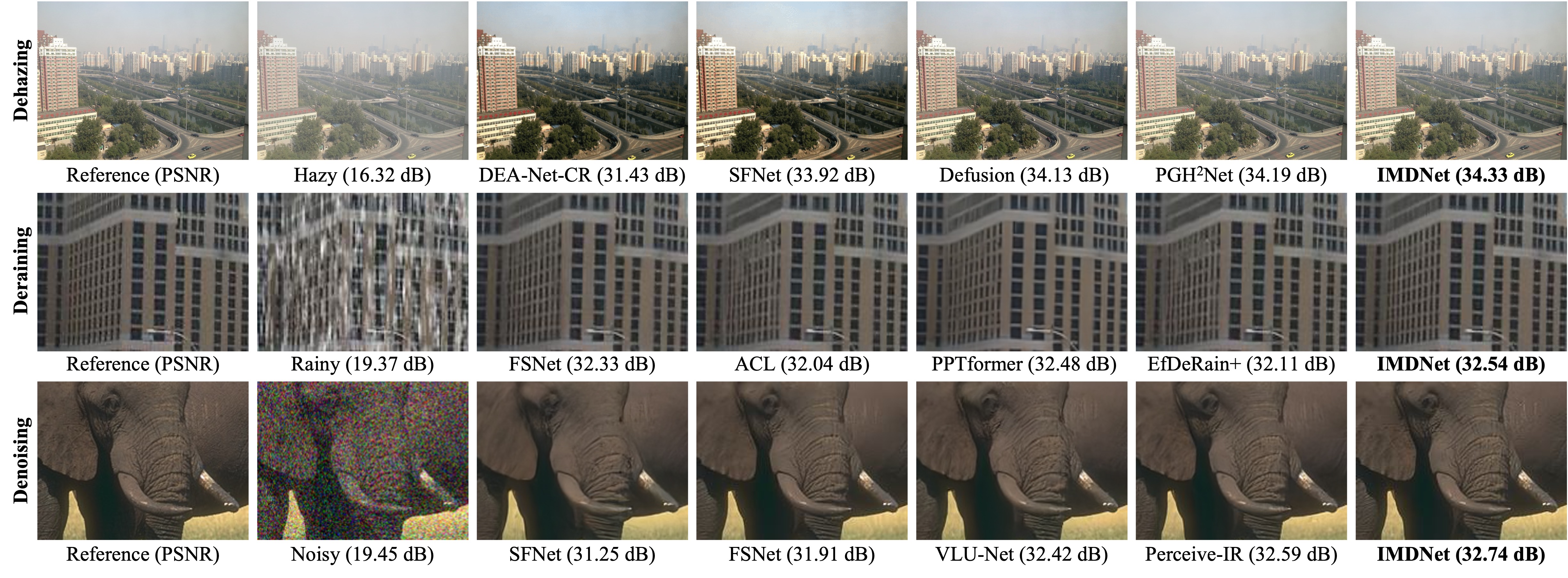}}
	\caption{Qualitative results under the SDIR experimental setup. Our IMDNet recovers finer details in the reconstructed images.}
 \label{fig:sig}
\end{figure*}

\section{Experiments}
\label{sec:exp}
In this section, we first describe the experimental setup, followed by qualitative and quantitative comparison results. Finally, we present ablation studies to validate the effectiveness of our approach. \textbf{Due to page limits, more experiments we show in the supplementary material.}

\subsection{Experimental Setup}
We conducted experiments under both MDIR and SDIR settings. \textbf{Datasets.} For MDIR, we use the dataset proposed by~\cite{FDTANetgao2025frequency}. For SDIR, the model is trained on clean-rain image pairs from multiple datasets~\cite{Rain100,Test100,8099669,7780668} and evaluated on various test sets, including Rain100H~\cite{Rain100}, Rain100L~\cite{Rain100}, Test100~\cite{Test100}, and Test1200~\cite{DIDMDN} for image deraining. For image denoising, images are collected from DIV2K~\cite{DIK}, Flickr2K~\cite{lim2017enhanced}, BSD500~\cite{BSD500}, and WED~\cite{ma2016waterloo}, with additive white Gaussian noise applied at levels from 0 to 50. The model is evaluated on CBSD68~\cite{BSD68}, Urban100~\cite{urban100}, and Kodak24~\cite{kodak}. For image dehazing, we use images from the RESIDE dataset~\citep{RESIDEli2018benchmarking} and evaluate on its SOTS subset~\citep{RESIDEli2018benchmarking}.

\textbf{Training details.} We train our models using the Adam optimizer~\cite{2014Adam} with parameters $\beta_1 = 0.9$ and $\beta_2 = 0.999$. The initial learning rate is set to $2 \times 10^{-4}$ and gradually decays to $1 \times 10^{-7}$ following a cosine annealing schedule~\cite{2016SGDR}. Training patches of size $256 \times 256$ are sampled with a batch size of 32 for $4 \times 10^5$ iterations. Data augmentation includes horizontal and vertical flips. For fair comparisons, all deep learning-based methods are fine-tuned or retrained using the parameter settings reported in their respective papers.

\subsection{Experimental Results}

\subsubsection{MDIR Setting}
We evaluate our IMDNet across seven combinations of degradation types, encompassing all permutations of haze, rain, and noise.
Table~\ref{tb:mix} reports the quantitative comparison results. On average across all  tasks, our method achieves a 1.03 dB improvement over the second-best approach, Defusion~\cite{DefusionLuo_2025_CVPR}. Compared with previous MDIR methods, Ref-IRT~\cite{REF-IRT} and FDTANet~\cite{FDTANetgao2025frequency}, our model shows substantial gains of 2.60 dB and 3.44 dB, respectively.

To further validate the adaptability of our approach to various degradation types, we conduct experiments under multiple degradation combinations using the same training dataset. The results demonstrate that IMDNet consistently achieves near-optimal or superior performance across diverse degradation scenarios.
Specifically, our model outperforms competing methods by 0.97 dB on the H+R+N combination compared to Perceive-IR~\cite{Perceive-IR10990319}, by 0.88 dB on H+R compared to Ref-IRT~\cite{REF-IRT}, and by 1.14 dB on N compared to Defusion~\cite{DefusionLuo_2025_CVPR}. Although IMDNet does not achieve the top performance on H+N, R+N, H, and R, the differences are marginal, within 0.06 dB.
As shown in Figure~\ref{fig:mix}, our model produces restored images that are sharper and visually closer to the ground truth than  others.

\subsubsection{SDIR Setting}
To verify that the proposed IMDNet is effective not only for MDIR but also performs well on SDIR, we conduct experiments on three representative image restoration tasks: image dehazing, image deraining, and image denoising. Figure~\ref{fig:sig} illustrates that the images restored by our IMDNet effectively reduce color distortion compared to other state-of-the-art methods. Moreover, our approach is able to reconstruct finer and sharper details in the degraded regions.

\begin{table}
    \centering
        \caption{Image dehazing results.}
    \label{tab:sot}
    \resizebox{\linewidth}{!}{
    \begin{tabular}{c|cccc}
    \hline
    \multicolumn{1}{c|}{} & \multicolumn{2}{c}{SOTS-Indoor}  & \multicolumn{2}{c}{SOTS-Outdoor} 
    \\
   Methods & PSNR $\uparrow$ & SSIM $\uparrow$  & PSNR $\uparrow$ & SSIM $\uparrow$ 
   \\
   \hline
   \hline
SFNet~\cite{SFNet} & 41.24 &\textbf{0.996} &\underline{40.05} &\textbf{0.996} 
\\
IRXNext~\cite{IRNeXt}&41.21 &\textbf{0.996} &39.18 &\textbf{0.996}       
\\
DEA-Net-CR~\cite{deanetchen2024dea} & 41.31&\underline{0.995} & 36.59 & 0.990 
\\
 Defusion~\cite{DefusionLuo_2025_CVPR} & 41.65&\underline{0.995} & 37.41 & \underline{0.993}
\\
PGH$^2$Net~\cite{PGH2Netisu2025prior}&\underline{41.70} &\textbf{0.996}&37.52 &0.989
         \\
         \hline
        \textbf{ IMDNet(Ours)} &\textbf{42.17}	&\textbf{0.996}	&\textbf{40.64}	&0.992
         \\
         \hline
    \end{tabular}}
\end{table}

\textbf{Image Dehazing.} We report the quantitative results of different image dehazing methods in Table~\ref{tab:sot}. Overall, our IMDNet consistently achieves superior performance across both indoor and outdoor scenarios. Specifically, on the indoor dataset SOT-indoor, IMDNet surpasses the previous best-performing method PGH$^2$Net~\cite{PGH2Netisu2025prior} by 0.47 dB in PSNR. On the outdoor dataset SOT-outdoor, it further achieves a 0.59 dB improvement over the previous state-of-the-art method SFNet~\cite{SFNet}.

\begin{table*}
\centering
\caption{Image deraining results.}
\label{tb:derain}
    \resizebox{\linewidth}{!}{
\begin{tabular}{c|cccccccc||cc}
    \hline
    \multicolumn{1}{c|}{} & \multicolumn{2}{c}{Test100~\citep{Test100}}  & \multicolumn{2}{c}{Test1200~\citep{MSPFN}} & \multicolumn{2}{c}{Rain100H~\citep{Rain100}} & \multicolumn{2}{c||}{Rain100L~\citep{Rain100}} & \multicolumn{2}{c}{Average} 
    \\
   Methods &PSNR $\uparrow$ &  SSIM $\uparrow$  & PSNR $\uparrow$ & SSIM $\uparrow$ &PSNR $\uparrow$ &SSIM $\uparrow$ & PSNR $\uparrow$&SSIM $\uparrow$ &PSNR $\uparrow$ & SSIM $\uparrow$
    \\
    \hline\hline
    FSNet~\citep{FSNet} &31.05&\underline{0.919} &33.08&0.916 & 31.77&0.906 &38.00 & 0.972 &33.48 &0.928
    \\
    MHNet~\citep{gao2025mixed} &31.25 &0.901 &\underline{33.45} &0.925 &31.08 &0.899 &\underline{40.04} &\textbf{0.985} &33.96 &0.928
     \\
     PPTformer~\cite{pptformerwang2025intra} & 31.48 & \textbf{0.922} & 33.39 & 0.911 &  31.77 & 0.907
     & 39.33 & \underline{0.983} & 33.99 & 0.931
     \\
 ACL~\cite{aclgu2025acl} &\underline{31.51} & 0.914 & 33.27 &\underline{0.928}  & 32.22 & \underline{0.920} & 39.18 &\underline{0.983}  & 34.05 & 0.936
     \\
     EfDeRain+~\cite{efderainguo2025efficientderain+} &31.10 &0.911 &33.12 & 0.925 & \textbf{34.57} &\textbf{0.957} & 39.03 & 0.972 &\underline{34.46} & \textbf{0.941}
     \\
      \hline
      \textbf{IMDNet(Ours)}  & \textbf{32.02}	&\textbf{0.922}	&\textbf{34.46}	&\textbf{0.939}	&\underline{32.42}	&0.914	&\textbf{40.10}	&\textbf{0.985} &\textbf{34.75}	&\underline{0.940}
    \\
    \hline
\end{tabular}}
\end{table*}

\textbf{Image Deraining.} Following the protocol in prior work~\cite{gao2025mixed}, we evaluate PSNR and SSIM metrics on the Y channel of the YCbCr color space for the image deraining task. As shown in Table~\ref{tb:derain}, our method consistently achieves comparable or superior results to existing approaches across all four benchmark datasets. In particular, IMDNet delivers an average PSNR improvement of 0.29 dB over the best-performing method EfDeRain+~\cite{efderainguo2025efficientderain+}. Moreover, on the Test100 dataset~\cite{Test100}, our model achieves a notable 0.52 dB gain compared to the previous leading method ACL~\cite{aclgu2025acl}, highlighting the strong deraining capability of our approach.

\begin{table}
    \centering
    \caption{Image denoising results.}
    \label{tab:denoisec}
    \resizebox{\linewidth}{!}{
    \begin{tabular}{c|ccccccccc}
    \hline
      & \multicolumn{3}{c}{CBSD68~\cite{BSD68}}  & \multicolumn{3}{c}{Kodak24~\cite{kodak}} & \multicolumn{3}{c}{Urban100~\cite{urban100}} 
    \\
   Methods  & 15  & 25 & 50 & 15  & 25 & 50 & 15  & 25 & 50
   \\
   \hline
   \hline
SFNet~\cite{SFNet} & 34.09 & 31.49 & 28.02 & 34.93 & 32.42 & 29.25&34.19 &32.01 & 29.03
\\
FSNet~\cite{FSNet} & 34.11 & 31.51 & 28.01 & \underline{34.95} & 32.42 & \underline{29.27}&34.15 & 32.04 & 29.15
\\
Perceive-IR~\cite{Perceive-IR10990319} &\underline{34.38} &\underline{31.74} &\underline{28.53} &34.84 &\underline{32.50} &29.16 &34.86 &\underline{32.55} &\underline{29.42}
\\
VLU-Net~\cite{VLUNetZeng_2025_CVPR}& 34.35 &31.72 &28.46 &- &- &-&\textbf{34.92} &\textbf{32.71} &\textbf{29.61}
\\
\hline
\textbf{IMDNet(Ours)} & \textbf{34.77}	&\textbf{32.14}	&	\textbf{29.17}	&	\textbf{35.55}	&	\textbf{33.01}	&	\textbf{30.09}	&	\underline{34.87}	&	32.34	&	29.39

       \\
    \hline
    \end{tabular}}
\end{table}

\textbf{Image Denoising.} Table~\ref{tab:denoisec} presents the performance of IMDNet model under different noise levels (15, 25, and 50). Our approach consistently delivers strong results across multiple datasets and noise intensities. In particular, at the challenging noise level of 50 on the CBSD68 dataset~\cite{BSD68}, IMDNet surpasses Perceive-IR~\cite{Perceive-IR10990319} by 0.64 dB. Similarly, for the same noise level on the Kodak24 dataset~\cite{kodak}, our method achieves a 0.82 dB improvement over the previous best-performing model FSNet~\cite{FSNet}. Although IMDNet does not reach the top performance on the Urban100 dataset~\cite{urban100}, the performance gap remains minimal.

\subsection{Ablation Studies}
We conduct ablation studies under the MDIR setting to demonstrate the effectiveness of our modules. We use the four-scale NAFNet~\cite{chen2022simple} as the baseline network and perform a step-by-step ablation study by successively integrating the proposed modules.

\begin{table}
    \centering
    \caption{Ablation study on individual components of IMDNet.}
    \label{tab:abl1}
       \resizebox{\linewidth}{!}{
    \begin{tabular}{ccc}
    \hline
         Method&  PSNR &$\triangle$ PSNR 
         \\
         \hline
         Baseline & 27.01 & -
         \\
         replace with DIDBlock& 27.64 & +0.63
        \\
         replace with TABlock& 28.55&  +1.54 
        \\
        replace with DIDBlock and TABlock & 30.91& +3.90
         \\
         replace with DIDBlock and TABlock + FBlock& 31.19& +4.18
         \\
         \hline
    \end{tabular}}
\end{table}

\subsubsection{Effectiveness of Each Module}
Table~\ref{tab:abl1} shows that the baseline model achieves a PSNR of 27.01 dB. When the encoder incorporates the proposed DIDBlock, the model effectively captures both spatial and frequency domain features, while reducing degradation information within the skip-connected features. This leads to more accurate information propagation and a 0.63 dB performance gain. However, the DIDBlock alone still lacks sufficient adaptability to handle diverse degradation types. When only the decoder is replaced with the TABlock, the model gains adaptive capability to distinguish and process different degradations. Yet, because its input primarily comes from the encoder output without fully decoupled degradation information, its recognition accuracy remains limited, resulting in only a 1.54 dB improvement. By jointly employing both DIDBlock and TABlock, the model achieves more precise degradation decoupling and information propagation through skip connections, leading to a significant 3.90 dB performance increase. Finally, with the integration of the FBlock to fuse degradation information across multiple scales, the model’s performance is further enhanced, reaching 31.19 dB.

\begin{table}
    \centering
    \caption{Plug-and-play ablation experiments.}
    \label{tab:ablplug}
    
    \begin{tabular}{ccc}
    \hline
         Method&  PSNR &$\triangle$ PSNR 
         \\
         \hline
         PromptIR~\cite{potlapalli2023promptir}& 27.54 & -
         \\
         $\triangle$ PromptIR~\cite{potlapalli2023promptir}& 30.88 & +3.34
        \\
        \hline
         VLU-Net~\cite{VLUNetZeng_2025_CVPR}& 29.35 & - 
        \\
         $\triangle$  VLU-Net~\cite{VLUNetZeng_2025_CVPR} & 31.13& +1.78
        
         \\
         \hline
    \end{tabular}
\end{table}

\subsubsection{Generality of the Overall Idea}
To further validate the  generality of our overall idea of decoupling degradation ingredient representations to guide path selection, we modify the encoder and decoder of existing methods. In the encoder, instead of changing its structure, we append a degradation information decoupling module after the original feature extraction stage. In the decoder, we adjust its feedforward network following the TABlock design. As shown in Table~\ref{tab:ablplug}, incorporating our proposed idea leads to significant performance gains, with PromptIR~\cite{potlapalli2023promptir} and VLU-Net~\cite{VLUNetZeng_2025_CVPR} achieving improvements of 3.34 dB and 1.78 dB, respectively.

%% file: sec/5_con.tex
\section{Conclusion}
\label{sec:con}

In this work, we propose IMDNet, an adaptive network for multi-degradation image restoration that addresses the challenges of images with coexisting degradations. By leveraging decoupled representations of degradation components, IMDNet effectively guides the selection of optimal restoration paths. The encoder’s degradation ingredient decoupling block (DIDBlock) separates degradation factors across spatial and frequency domains, enhancing recognition and ensuring independent feature representations. The fusion block (FBlock) integrates multi-level degradation information, while the decoder’s task adaptation block (TABlock) dynamically activates or fuses functional branches based on the multi-degradation representation. Extensive experiments demonstrate that IMDNet excels in MDIR while remaining highly competitive on SDIR.

%% file: sec/X_suppl.tex
\clearpage
\setcounter{page}{1}
\maketitlesupplementary

\section{Overview}
\label{sec:over}
The Appendix is composed of:

Dataset~\ref{sec:data}

More Experiments~\ref{sec:mas}

Additional Visual Results~\ref{sec:Visual}

\section{Dataset}
\label{sec:data}

\subsection{MDIR}
Following~\cite{FDTANetgao2025frequency}, we conduct comparative experiments on a dataset containing various combinations of degradations, including haze, rain, and noise. The dataset consists of 18,000 natural images collected from several public image restoration datasets~\citep{RESIDEli2018benchmarking,Rain100,Test100,BSDmartin2001database}.
Among them, 13,000 images are randomly selected for training, where haze, rain, and noise degradations are synthetically introduced with intensity levels of [0,150], [0,300], and [0,50], respectively. Among the remaining,  500 images are randomly selected for testing. As summarized in Table~\ref{tab:testco}, the test set covers seven distinct degradation combinations, encompassing all permutations of haze, rain, and noise.
For example, in the “haze + rain + noise” setting, the same degradation processes as in training are applied, while in the “haze + rain” case, the random noise addition is omitted, with other settings kept identical. The remaining configurations are constructed in a similar fashion.

\subsection{SDIR}
\subsubsection{Image Deraining}
Following the experimental setups of recent state-of-the-art methods for image deraining~\cite{Zamir2021MPRNet,FSNet}, we train our model using 13,712 clean-rain image pairs collected from multiple datasets~\cite{Rain100,Test100,8099669,7780668}. Using the trained IMDNet, we conduct evaluations on various test sets, including Rain100H~\cite{Rain100}, Rain100L~\cite{Rain100}, Test100~\cite{Test100}, and Test1200~\cite{DIDMDN}.

\subsubsection{Image Denoising}
In line with the methodology of~\cite{Zamir2021Restormer}, we train a single IMDNet model capable of handling various noise levels on a composite dataset. This dataset comprises 800 images from DIV2K~\cite{DIK}, 2,650 images from Flickr2K~\cite{lim2017enhanced}, 400 images from BSD500~\cite{BSD500}, and 4,744 images from WED~\cite{ma2016waterloo}. The noisy images are generated by adding additive white Gaussian noise with a random noise level $\sigma$ chosen from the set [15, 25, 50] to the clean images. Testing is conducted on the CBSD68~\cite{BSD68}, Urban100~\cite{urban100}, and Kodak24~\cite{kodak} benchmark datasets.

\subsubsection{Image Dehazing} 
We assess the performance of our IMDNet on the daytime datasets encompass synthetic data from RESIDE~\cite{RESIDEli2018benchmarking}. The RESIDE dataset contains two training subsets: the indoor training set (ITS) and the outdoor training set (OTS), along with a synthetic objective testing set (SOTS). The ITS consists of 13,990 hazy images generated from 1,399 sharp images, while the OTS comprises 313,950 hazy images derived from 8,970 clean images.
Our model is trained separately on the ITS and OTS datasets, and subsequently tested on their corresponding test sets, namely SOTS-Indoor and SOTS-Outdoor, each comprising 500 paired images. 

\subsubsection{Image Desnowing}
For desnowing evaluation, we utilize three datasets: Snow100K~\cite{desnownet}, SRRS~\cite{JSTASRchen2020jstasr}, and CSD~\cite{HDCW-Netchen2021all}. Snow100K~\cite{desnownet} comprises 50,000 image pairs for training and 50,000 for evaluation. SRRS~\cite{JSTASRchen2020jstasr} contains 15,005 image pairs for training and 15,005 for evaluation. CSD~\cite{HDCW-Netchen2021all} consists of 8,000 image pairs for training and 2,000 for evaluation.
To maintain consistency with the training strategy of the previous algorithm~\cite{FSNet}, we randomly sample 2,500 image pairs from the training set for training and 2,000 images from the testing set for evaluation. 

\begin{table}
    \centering
    \caption{Test set generation by multi-degradation types.}
    \label{tab:testco}
    \resizebox{\linewidth}{!}{
    \begin{tabular}{cccc}
        \hline
      Multi-degradation & Haze level & Rain level & Noise level
      \\
      \hline
      Haze + Rain + Noise & [0,150] & [0,300] & [0,50]
      \\
       Haze + Rain  & [0,150] & [0,300] & [0]
       \\
       Haze  + Noise & [0,150] & [0] & [0,50]
        \\
          Rain  + Noise & [0] & [0,300] & [0,50]
        \\
          Haze  & [0,150] & [0] & [0]
        \\
          Rain & [0] & [0,300] & [0]
        \\
         Noise & [0] & [0] & [0,50]
        \\
        \hline
    \end{tabular}} 
\end{table}

\begin{figure*} % use float package if you want it here
    \centerline{\includegraphics[width=1\textwidth]{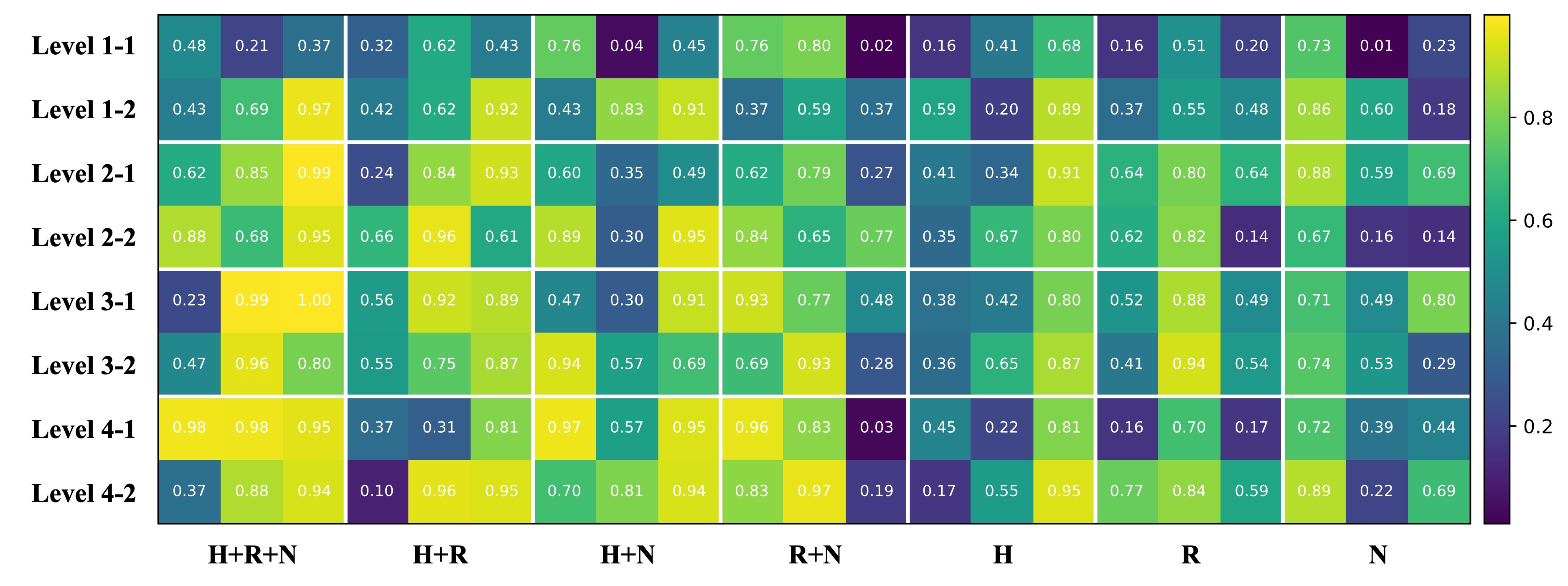}}
	\caption{Heat-maps for the activation weight of TABlock.}
 \label{fig:taabl}
\end{figure*}

\section{More Experiments}
\label{sec:mas}

\subsection{More Experimental Results}
To further validate the effectiveness of our approach, we perform image desnowing experiments under the SSIR setting.
\subsubsection{Image Desnowing}
We conduct a comprehensive comparison of desnowing performance across three benchmark datasets. As summarized in Table~\ref{tab:snow}, our proposed IMDNet consistently surpasses existing state-of-the-art methods by a significant margin. In particular, IMDNet achieves an impressive 2.07 dB gain on the SRRS dataset~\cite{JSTASRchen2020jstasr} and a 1.15 dB improvement on the recently introduced CSD dataset~\cite{HDCW-Netchen2021all} when compared with IRNeXt~\cite{IRNeXt}. These results highlight the model’s superior capability in handling diverse snow patterns and background complexities. Furthermore, relative to MSP-Former~\cite{mspformer10095605}, a task-specific architecture tailored for image desnowing, IMDNet exhibits substantial quantitative gains of 4.69 dB, 3.22 dB, and 1.28 dB on the Snow100K, SRRS, and CSD datasets, respectively. This demonstrates that our adaptive multi-degradation design not only generalizes effectively to snow removal tasks but also achieves state-of-the-art restoration quality across different data distributions.

\begin{table}
    \centering
        \caption{Quantitative comparisons with other image  desnowing methods.}
    \label{tab:snow}
    \resizebox{\linewidth}{!}{
    \begin{tabular}{c|cccccc}
    \hline
    \multicolumn{1}{c|}{} & \multicolumn{2}{c}{CSD}  & \multicolumn{2}{c}{SRRS} & \multicolumn{2}{c}{Snow100K}
    \\
   Methods & PSNR $\uparrow$ & SSIM $\uparrow$  & PSNR $\uparrow$ & SSIM $\uparrow$ & PSNR $\uparrow$ & SSIM $\uparrow$
   \\
   \hline
   \hline
 NAFNet~\cite{chen2022simple} &33.13 &0.96 &29.72 &0.94 &32.41 &\underline{0.95}
\\
FocalNet~\cite{focalnetcui2023focal} &37.18 &\textbf{0.99} &31.34 &\textbf{0.98} &33.53 &\underline{0.95}
\\
MSP-Former~\cite{mspformer10095605} & 33.75 &0.96 &30.76 &0.95 &33.43& \textbf{0.96}
\\
IRNeXt~\cite{IRNeXt} &\underline{37.29} &\textbf{0.99} &\underline{31.91} &\textbf{0.98} &\underline{33.61} &\underline{0.95}
\\
\hline
\textbf{IMDNet(Ours)} &\textbf{38.44}	&\underline{0.98}	&\textbf{33.98}	&\underline{0.96}	&\textbf{34.71}	&\textbf{0.96}

         \\
         \hline
    \end{tabular}}
\end{table}

\subsection{More Ablation Studies}

\subsubsection{Effectiveness of TABlock}
To illustrate the adaptive capability of our TABlock, we show that it can dynamically activate or fuse multiple functional branches based on the decoupled degradation information, thereby flexibly selecting the optimal restoration path and handling various combinations of degraded images. We visualize the activation weight vector for each branch to provide insight into this process. As shown in Figure~\ref{fig:taabl}, the activation values vary across different degradation combinations. Specifically, when all degradation types are present, all branches are activated and fused. In contrast, for other combinations, certain branches are either not selected or assigned low weights. This clearly demonstrates TABlock’s ability to adaptively adjust its processing according to the specific degradation characteristics of the input image.

\begin{figure*} % use float package if you want it here
    \centerline{\includegraphics[width=1\textwidth]{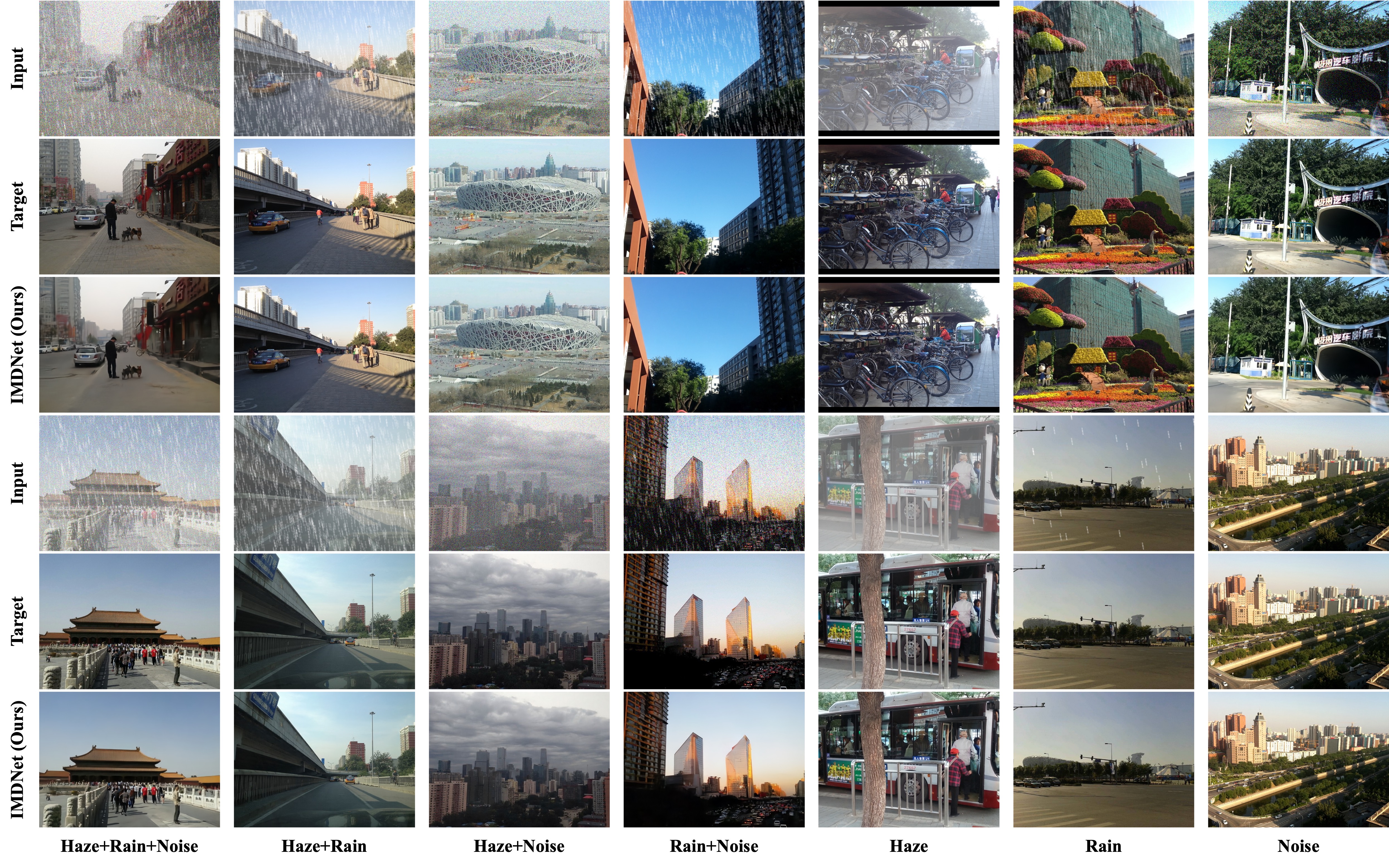}}
	\caption{Image restoration results across various combinations of degradation types. The top row displays the degraded inputs, the middle row presents the ground truth images, and the bottom row shows the corresponding restoration results generated by IMDNet.}
 \label{mix_v}
\end{figure*}

\subsubsection{Design Choices of Skip Connection}

To further demonstrate the effectiveness of employing our decoupled features $CF_i$ for skip connections, we conduct a comparative experiment using the directly encoded features $E_i$ from the encoder. The results, summarized in Table~\ref{tab:dcsc}, clearly show that our approach achieves superior performance. This confirms that the proposed DIDBlock effectively separates degradation-related information from clean feature representations, ensuring that the skip connections transmit more reliable and noise-free information to the decoder, thereby enhancing the overall restoration quality.

\begin{table}
    \centering
    \caption{Design Choices of Skip Connection.}
    \label{tab:dcsc}
       \resizebox{\linewidth}{!}{
    \begin{tabular}{ccc}
    \hline
         Method&  PSNR &$\triangle$ PSNR 
         \\
         \hline
         Encoder feature $E_i$ & 30.98 & -
         \\
         Decoupled clean feature $CF_i$& 31.19 & +0.21
         \\
         \hline
    \end{tabular}}
\end{table}

\subsubsection{Effectiveness of Decoupling Loss}
To enable accurate feature decoupling in the encoder, the skip-connection feature $CF$ is constrained to contain minimal degradation information, while the branch-selection feature 
$DI$ is encouraged to primarily capture degradation-specific cues. We impose this separation through a cosine similarity loss that maximizes the distinction between $CF$ and $DI$. To evaluate the effectiveness of this decoupling loss, we conducted corresponding experiments, as shown in Table~\ref{tab:deloss}. The results demonstrate that incorporating this loss helps the model better disentangle the respective features, leading to improved restoration performance under multiple degradation conditions.

\begin{table}
    \centering
    \caption{Effectiveness of Decoupling Loss.}
    \label{tab:deloss}
    \begin{tabular}{ccc}
    \hline
         Method&  PSNR &$\triangle$ PSNR 
         \\
         \hline
         w/o $L_{d}$& 30.66 & -
         \\
         w/ $L_{d}$ & 31.19 & +0.53
         \\
         \hline
    \end{tabular}
\end{table}

\begin{table}
    \caption{The effect of different degradation information aggregation methods on overall performance.}
    \label{tab:ablfa}
    \centering
     
    \begin{tabular}{ccccc}
    \hline
         Modules&  Sum&  Concatenation  & FBlock
         \\
         \hline
         PSNR&  31.01  &31.07  &31.19
         \\
         FLOPs(G) &23.15 & 24.22 &23.15
         \\
         \hline
    \end{tabular}
\end{table}

\subsubsection{Design Choices for FBlock}
To evaluate the effectiveness of the proposed FBlock, we compare it with alternative aggregation strategies, including simple summation and concatenation. As reported in Table~\ref{tab:ablfa}, our FBlock achieves consistently better performance, demonstrating its ability to effectively integrate degradation features across multiple hierarchical levels. Notably, this improvement is achieved without adding any extra computational overhead.

\begin{table}
    \centering
    \caption{The evaluation of model computational complexity. }
    \label{tab:computational}
     \resizebox{\linewidth}{!}{
    \begin{tabular}{ccccc}
    \hline
         Method& Time(s) & Params(M)  & PSNR & SSIM
         \\
         \hline\hline
         VLU-Net~\cite{VLUNetZeng_2025_CVPR} &0.743 &35  &29.35 & 0.842
         \\
         PromptIR~\cite{potlapalli2023promptir} &1.012 & 33 & 27.54 & 0.819
         \\
         Perceive-IR~\cite{Perceive-IR10990319} &0.682 &45 &30.22 & 0.894 
         \\
         \hline
        \textbf{IMDNet(Ours)} &0.352 &24 &31.19 & 0.910
        \\
         \hline
    \end{tabular}}
\end{table}

\subsection{Resource Efficient}
We further analyze the computational efficiency of IMDNet by comparing its runtime and parameter count with recent state-of-the-art methods. As shown in Table~\ref{tab:computational}, IMDNet not only achieves leading restoration performance but also significantly reduces computational demand. In particular, it surpasses the previous best method, Perceive-IR~\cite{Perceive-IR10990319}, by 0.2 dB while requiring only 51.6\% of its inference time.

\section{Additional Visual Results}
\label{sec:Visual}

Figure~\ref{mix_v} illustrates the visual results of our method when handling different combinations of degradation types. In each case, the upper row presents the input degraded images affected by multiple degradations, while the lower row displays the corresponding restored outputs produced by IMDNet. As observed, our model effectively removes various types of degradations and consistently reconstructs images with sharp details, natural textures, and visually pleasing quality across all scenarios.

%% file: main.bib
@String(CVPR= {IEEE Conf. Comput. Vis. Pattern Recog.})

@String(ICCV= {Int. Conf. Comput. Vis.})

@String(ECCV= {Eur. Conf. Comput. Vis.})

@String(TIP  = {IEEE Trans. Image Process.})

@String(ICASSP=	{ICASSP})

@String(ACCV  = {ACCV})

@String(ICLR = {Int. Conf. Learn. Represent.})

@String(AAAI = {AAAI})

@String(CVPRW= {IEEE Conf. Comput. Vis. Pattern Recog. Worksh.})

@String(CVPR  = {CVPR})

@String(ICCV  = {ICCV})

@String(ECCV  = {ECCV})

@String(TIP   = {IEEE TIP})

@String(TCSVT = {IEEE TCSVT})

@String(ICLR  = {ICLR})

@String(CVPRW= {CVPRW})

@article{2011Single,
  title={Single Image Haze Removal Using Dark Channel Prior.},
  author={ He, Kaiming  and  Sun, Jian  and  Tang, Xiaoou },
  journal={TPAMI},
  year={2011},
}

@inproceedings{all_conli2022all,
  title={All-in-one image restoration for unknown corruption},
  author={Li, Boyun and Liu, Xiao and Hu, Peng and Wu, Zhongqin and Lv, Jiancheng and Peng, Xi},
  booktitle={CVPR},
  pages={17452--17462},
  year={2022}
}

@article{RESIDEli2018benchmarking,
  title={Benchmarking single-image dehazing and beyond},
  author={Li, Boyi and Ren, Wenqi and Fu, Dengpan and Tao, Dacheng and Feng, Dan and Zeng, Wenjun and Wang, Zhangyang},
  journal={TIP},
  volume={28},
  number={1},
  pages={492--505},
  year={2018},
  publisher={IEEE}
}

@ARTICLE{u2former,
  author={Feng, Xin and Ji, Haobo and Pei, Wenjie and Li, Jinxing and Lu, Guangming and Zhang, David},
  journal={TCSVT}, 
  title={U2-Former: Nested U-shaped Transformer for Image Restoration via Multi-view Contrastive Learning}, 
  year={2023},
  volume={},
  number={},
  pages={1-1},
}

@inproceedings{Zamir2021MPRNet,
    title={Multi-Stage Progressive Image Restoration},
    author={Syed Waqas Zamir and Aditya Arora and Salman Khan and Munawar Hayat
            and Fahad Shahbaz Khan and Ming-Hsuan Yang and Ling Shao},
    booktitle={CVPR},
    year={2021}
}

@inproceedings{Zamir2021Restormer,
    title={Restormer: Efficient Transformer for High-Resolution Image Restoration}, 
    author={Syed Waqas Zamir and Aditya Arora and Salman Khan and Munawar Hayat 
            and Fahad Shahbaz Khan and Ming-Hsuan Yang},
    booktitle={CVPR},
    year={2022}
}

@article{chen2022simple,
  title={Simple Baselines for Image Restoration},
  author={Chen, Liangyu and Chu, Xiaojie and Zhang, Xiangyu and Sun, Jian},
  journal={ECCV},
  year={2022}
}

@article{2014Adam,
  title={Adam: A Method for Stochastic Optimization},
  author={ Kingma, D.  and  Ba, J. },
  journal={Computer Science},
  year={2014},
}

@article{2016SGDR,
  title={SGDR: Stochastic Gradient Descent with Warm Restarts},
  author={ Loshchilov, I.  and  Hutter, F. },
  year={2016},
}

@article{MSPFN,
  title={Multi-Scale Progressive Fusion Network for Single Image Deraining},
  author={ Jiang, Kui  and  Wang, Zhongyuan  and  Yi, Peng  and  Chen, Chen  and  Huang, Baojin  and  Luo, Yimin  and  Ma, Jiayi  and  Jiang, Junjun },
journal={CVPR},
  year={2020},
}

@article{DIDMDN,
  title={Density-Aware Single Image De-raining Using a Multi-stream Dense Network},
  author={He Zhang and Vishal M. Patel},
  journal={CVPR},
  year={2018},
  pages={695-704}
}

@article{Rain100,
  title={Deep Joint Rain Detection and Removal from a Single Image},
  author={Wenhan Yang and Robby T. Tan and Jiashi Feng and Jiaying Liu and Zongming Guo and Shuicheng Yan},
  journal={CVPR},
  year={2016},
  pages={1685-1694}
}

@article{Test100,
  title={Image De-Raining Using a Conditional Generative Adversarial Network},
  author={He Zhang and Vishwanath A. Sindagi and Vishal M. Patel},
  journal={TCSVT},
  year={2017},
  volume={30},
  pages={3943-3956}
}

@inproceedings{BSDmartin2001database,
  title={A database of human segmented natural images and its application to evaluating segmentation algorithms and measuring ecological statistics},
  author={Martin, David and Fowlkes, Charless and Tal, Doron and Malik, Jitendra},
  booktitle={ICCV},
  volume={2},
  pages={416--423},
  year={2001},
  organization={IEEE}
}

@inproceedings{SFNet,
  title={Selective Frequency Network for Image Restoration},
  author={Cui, Yuning and Tao, Yi and Bing, Zhenshan and Ren, Wenqi and Gao, Xinwei and Cao, Xiaochun and Huang, Kai and Knoll, Alois},
  booktitle={ICLR},
  year={2023}
}

@ARTICLE{FSNet,
  author={Cui, Yuning and Ren, Wenqi and Cao, Xiaochun and Knoll, Alois},
  journal={TPAMI}, 
  title={Image Restoration Via Frequency Selection}, 
  year={2023},
  volume={},
  number={},
  pages={1-16},
}

@INPROCEEDINGS{tanet,
  author={Zhou, Jingyuan and Leong, Chaktou and Lin, Minyi and Liao, Wantong and Li, Congduan},
  booktitle={WACVW}, 
  title={Task Adaptive Network for Image Restoration with Combined Degradation Factors}, 
  year={2022},
  volume={},
  number={},
  pages={1-8}}

@inproceedings{AASO,
    Author = {M. Suganuma and X. Liu and T. Okatani},
    Title = {Attention-based Adaptive Selection of Operations for Image Restoration in the Presence of Unknown Combined Distortions},
    Booktitle = {CVPR},
    Year = {2019}
}

@article{potlapalli2023promptir,
title={PromptIR: Prompting for All-in-One Blind Image Restoration},
author={Potlapalli, Vaishnav and Zamir, Syed Waqas and Khan, Salman and Khan, Fahad Shahbaz},
journal={Advances in Neural Information Processing Systems (NeurIPS)},
year={2023}
}

@ARTICLE{10558778,
  author={Rong, Jianxiang and Huang, Hua and Li, Jia},
  journal={IEEE Transactions on Image Processing}, 
  title={IMU-Assisted Accurate Blur Kernel Re-Estimation in Non-Uniform Camera Shake Deblurring}, 
  year={2024},
  volume={33},
  number={},
  pages={3823-3838},
 }

@ARTICLE{BSD500,
  author={Arbeláez, Pablo and Maire, Michael and Fowlkes, Charless and Malik, Jitendra},
  journal={IEEE Transactions on Pattern Analysis and Machine Intelligence}, 
  title={Contour Detection and Hierarchical Image Segmentation}, 
  year={2011},
  volume={33},
  number={5},
  pages={898-916},
  doi={10.1109/TPAMI.2010.161}}

@article{ma2016waterloo,
  title={Waterloo exploration database: New challenges for image quality assessment models},
  author={Ma, Kede and Duanmu, Zhengfang and Wu, Qingbo and Wang, Zhou and Yong, Hongwei and Li, Hongliang and Zhang, Lei},
  journal={IEEE Transactions on Image Processing},
  volume={26},
  number={2},
  pages={1004--1016},
  year={2016},
  publisher={IEEE}
}

@ARTICLE{desnownet,
  author={Liu, Yun-Fu and Jaw, Da-Wei and Huang, Shih-Chia and Hwang, Jenq-Neng},
  journal={IEEE Transactions on Image Processing}, 
  title={DesnowNet: Context-Aware Deep Network for Snow Removal}, 
  year={2018},
  volume={27},
  number={6},
  pages={3064-3073},
  doi={10.1109/TIP.2018.2806202}
}

@INPROCEEDINGS{mspformer10095605,
  author={Chen, Sixiang and Ye, Tian and Liu, Yun and Liao, Taodong and Jiang, Jingxia and Chen, Erkang and Chen, Peng},
  booktitle={ICASSP 2023 - 2023 IEEE International Conference on Acoustics, Speech and Signal Processing (ICASSP)}, 
  title={MSP-Former: Multi-Scale Projection Transformer for Single Image Desnowing}, 
  year={2023},
  volume={},
  number={},
  pages={1-5},
  doi={10.1109/ICASSP49357.2023.10095605}}

@inproceedings{HDCW-Netchen2021all,
  title={All snow removed: Single image desnowing algorithm using hierarchical dual-tree complex wavelet representation and contradict channel loss},
  author={Chen, Wei-Ting and Fang, Hao-Yu and Hsieh, Cheng-Lin and Tsai, Cheng-Che and Chen, I and Ding, Jian-Jiun and Kuo, Sy-Yen and others},
  booktitle={Proceedings of the IEEE/CVF International Conference on Computer Vision},
  pages={4196--4205},
  year={2021}
}

@inproceedings{JSTASRchen2020jstasr,
  title={JSTASR: Joint size and transparency-aware snow removal algorithm based on modified partial convolution and veiling effect removal},
  author={Chen, Wei-Ting and Fang, Hao-Yu and Ding, Jian-Jiun and Tsai, Cheng-Che and Kuo, Sy-Yen},
  booktitle={Computer Vision--ECCV 2020: 16th European Conference, Glasgow, UK, August 23--28, 2020, Proceedings, Part XXI 16},
  pages={754--770},
  year={2020},
  organization={Springer}
}

@inproceedings{lim2017enhanced,
  title={Enhanced deep residual networks for single image super-resolution},
  author={Lim, Bee and Son, Sanghyun and Kim, Heewon and Nah, Seungjun and Mu Lee, Kyoung},
  booktitle={Proceedings of the IEEE conference on computer vision and pattern recognition workshops},
  pages={136--144},
  year={2017}
}

@INPROCEEDINGS{DIK,
  author={Agustsson, Eirikur and Timofte, Radu},
  booktitle={2017 IEEE Conference on Computer Vision and Pattern Recognition Workshops (CVPRW)}, 
  title={NTIRE 2017 Challenge on Single Image Super-Resolution: Dataset and Study}, 
  year={2017},
  volume={},
  number={},
  pages={1122-1131},
  doi={10.1109/CVPRW.2017.150}}

@INPROCEEDINGS{7780668,
  author={Li, Yu and Tan, Robby T. and Guo, Xiaojie and Lu, Jiangbo and Brown, Michael S.},
  booktitle={2016 IEEE Conference on Computer Vision and Pattern Recognition (CVPR)}, 
  title={Rain Streak Removal Using Layer Priors}, 
  year={2016},
  volume={},
  number={},
  pages={2736-2744},
  doi={10.1109/CVPR.2016.299}}

@INPROCEEDINGS{8099669,
  author={Fu, Xueyang and Huang, Jiabin and Zeng, Delu and Huang, Yue and Ding, Xinghao and Paisley, John},
  booktitle={2017 IEEE Conference on Computer Vision and Pattern Recognition (CVPR)}, 
  title={Removing Rain from Single Images via a Deep Detail Network}, 
  year={2017},
  volume={},
  number={},
  pages={1715-1723},
  doi={10.1109/CVPR.2017.186}}

@INPROCEEDINGS{OWN,
  author={Huang, Zihao and Li, Chao and Duan, Feng and Zhao, Qibin},
  booktitle={2021 International Joint Conference on Neural Networks (IJCNN)}, 
  title={Multi-distorted Image Restoration with Tensor 1 × 1 Convolutional Layer}, 
  year={2021},
  volume={},
  number={},
  pages={1-8},
 }

@INPROCEEDINGS{adarevD10656920,
  author={Mao, Xintian and Li, Qingli and Wang, Yan},
  booktitle={2024 IEEE/CVF Conference on Computer Vision and Pattern Recognition (CVPR)}, 
  title={AdaRevD: Adaptive Patch Exiting Reversible Decoder Pushes the Limit of Image Deblurring}, 
  year={2024},
  volume={},
  number={},
  pages={25681-25690},
}

@inproceedings{upid10.1145/3664647.3680560,
author = {Wen, Yuanbo and Gao, Tao and Chen, Ting},
title = {Unpaired Photo-realistic Image Deraining with Energy-informed Diffusion Model},
year = {2024},
booktitle = {Proceedings of the 32nd ACM International Conference on Multimedia},
pages = {360–369},
numpages = {10},
}

@ARTICLE{NDR10680296,
  author={Yao, Mingde and Xu, Ruikang and Guan, Yuanshen and Huang, Jie and Xiong, Zhiwei},
  journal={IEEE Transactions on Image Processing}, 
  title={Neural Degradation Representation Learning for All-in-One Image Restoration}, 
  year={2024},
  volume={33},
  number={},
  pages={5408-5423},
  }

@InProceedings{liu2024xyscannet,
   title={XYScanNet: An Interpretable State Space Model for Perceptual Image Deblurring},
  author={Liu, Hanzhou and Liu, Chengkai and Xu, Jiacong and Jiang, Peng and Lu, Mi},
    booktitle = {Proceedings of the Computer Vision and Pattern Recognition Conference (CVPR)},
    year      = {2025},
    pages     = {779-789}
}

@inproceedings{autodir10.1007/978-3-031-73661-2_19,
author = {Jiang, Yitong and Zhang, Zhaoyang and Xue, Tianfan and Gu, Jinwei},
title = {AutoDIR: Automatic All-in-One Image Restoration with Latent Diffusion},
year = {2024},
booktitle = {Computer Vision – ECCV 2024: 18th European Conference, Milan, Italy, September 29–October 4, 2024, Proceedings, Part XL},
pages = {340–359},
numpages = {20},

}

@inproceedings{guo2025mambairv2,
  title={Mambairv2: Attentive state space restoration},
  author={Guo, Hang and Guo, Yong and Zha, Yaohua and Zhang, Yulun and Li, Wenbo and Dai, Tao and Xia, Shu-Tao and Li, Yawei},
  booktitle={Proceedings of the Computer Vision and Pattern Recognition Conference},
  pages={28124--28133},
  year={2025}
}

@inproceedings{cui2025adair,
title={Ada{IR}: Adaptive All-in-One Image Restoration via Frequency Mining and Modulation},
author={Yuning Cui and Syed Waqas Zamir and Salman Khan and Alois Knoll and Mubarak Shah and Fahad Shahbaz Khan},
booktitle={The Thirteenth International Conference on Learning Representations},
year={2025}
}

@inproceedings{guo2025mambair,
  title={MambaIR: A simple baseline for image restoration with state-space model},
  author={Guo, Hang and Li, Jinmin and Dai, Tao and Ouyang, Zhihao and Ren, Xudong and Xia, Shu-Tao},
  booktitle={European Conference on Computer Vision},
  pages={222--241},
  year={2024}
}

@INPROCEEDINGS{diffunmix10656884,
  author={Zeng, Haijin and Cao, Jiezhang and Zhang, Kai and Chen, Yongyong and Luong, Hiep and Philips, Wilfried},
  booktitle={2024 IEEE/CVF Conference on Computer Vision and Pattern Recognition (CVPR)}, 
  title={Unmixing Diffusion for Self-Supervised Hyperspectral Image Denoising}, 
  year={2024},
  volume={},
  number={},
  pages={27820-27830},
}

@inproceedings{ALGgao2024learning,
  title={Learning enriched features via selective state spaces model for efficient image deblurring},
  author={Gao, Hu and Ma, Bowen and Zhang, Ying and Yang, Jingfan and Yang, Jing and Dang, Depeng},
  booktitle={Proceedings of the 32nd ACM International Conference on Multimedia},
  pages={710--718},
  year={2024}
}

@inproceedings{LSSRgao2024learning,
  title={Learning optimal combination patterns for lightweight stereo image super-resolution},
  author={Gao, Hu and Yang, Jing and Zhang, Ying and Yang, Jingfan and Ma, Bowen and Dang, Depeng},
  booktitle={Proceedings of the 32nd ACM International Conference on Multimedia},
  pages={5566--5574},
  year={2024}
}

@InProceedings{Kim_2020_ACCV,
    author    = {Kim, Sijin and Ahn, Namhyuk and Sohn, Kyung-Ah},
    title     = {Restoring Spatially-Heterogeneous Distortions using Mixture of Experts Network},
    booktitle = {Proceedings of the Asian Conference on Computer Vision (ACCV)},
    month     = {November},
    year      = {2020}
}

@INPROCEEDINGS{DIGNet,
  author={Shin, Wooksu and Ahn, Namhyuk and Moon, Jeong-Hyeon and Sohn, Kyung-Ah},
  booktitle={2022 IEEE/CVF Conference on Computer Vision and Pattern Recognition Workshops (CVPRW)}, 
  title={Exploiting Distortion Information for Multi-degraded Image Restoration}, 
  year={2022},
  volume={},
  number={},
  pages={536-545},
 }

@ARTICLE{REF-IRT,
  author={Zhang, Yi and Yang, Qixue and Chandler, Damon M. and Mou, Xuanqin},
  journal={IEEE Transactions on Image Processing}, 
  title={Reference-Based Multi-Stage Progressive Restoration for Multi-Degraded Images}, 
  year={2024},
  volume={33},
  number={},
  pages={4982-4997},
 }

@INPROCEEDINGS{urban100,
  author={Huang, Jia-Bin and Singh, Abhishek and Ahuja, Narendra},
  booktitle={2015 IEEE Conference on Computer Vision and Pattern Recognition (CVPR)}, 
  title={Single image super-resolution from transformed self-exemplars}, 
  year={2015},
  volume={},
  number={},
  pages={5197-5206},
 }

@INPROCEEDINGS{BSD68,
  author={Martin, D. and Fowlkes, C. and Tal, D. and Malik, J.},
  booktitle={Proceedings Eighth IEEE International Conference on Computer Vision. ICCV 2001}, 
  title={A database of human segmented natural images and its application to evaluating segmentation algorithms and measuring ecological statistics}, 
  year={2001},
  volume={2},
  number={},
  pages={416-423 vol.2},
}

@article{efderainguo2025efficientderain+,
  title={EfficientDeRain+: Learning Uncertainty-Aware Filtering via RainMix Augmentation for High-Efficiency Deraining},
  author={Guo, Qing and Qi, Hua and Sun, Jingyang and Juefei-Xu, Felix and Ma, Lei and Lin, Di and Feng, Wei and Wang, Song},
  journal={International Journal of Computer Vision},
  volume={133},
  number={4},
  pages={2111--2135},
  year={2025},
}

@inproceedings{aclgu2025acl,
  title={ACL: Activating Capability of Linear Attention for Image Restoration},
  author={Gu, Yubin and Meng, Yuan and Ji, Jiayi and Sun, Xiaoshuai},
  booktitle={Proceedings of the Computer Vision and Pattern Recognition Conference},
  pages={17913--17923},
  year={2025}
}

@inproceedings{pptformerwang2025intra,
  title={Intra and Inter Parser-Prompted Transformers for Effective Image Restoration},
  author={Wang, Cong and Pan, Jinshan and Wang, Liyan and Wang, Wei},
  booktitle={Proceedings of the AAAI Conference on Artificial Intelligence},
  volume={39},
  number={7},
  pages={7609--7618},
  year={2025}
}

@article{gao2025mixed,
  title={Mixed hierarchy network for image restoration},
  author={Gao, Hu and Zhang, Ying and Yang, Jing and Dang, Depeng},
  journal={Pattern Recognition},
  volume={161},
  pages={111313},
  year={2025},
}

@ARTICLE{Perceive-IR10990319,
  author={Zhang, Xu and Ma, Jiaqi and Wang, Guoli and Zhang, Qian and Zhang, Huan and Zhang, Lefei},
  journal={IEEE Transactions on Image Processing}, 
  title={Perceive-IR: Learning to Perceive Degradation Better for All-in-One Image Restoration}, 
  year={2025},
  volume={},
  number={},
  pages={1-1},
  }

@inproceedings{IRNeXt,
author = {Cui, Yuning and Ren, Wenqi and Yang, Sining and Cao, Xiaochun and Knoll, Alois},
title = {IRNeXt: Rethinking Convolutional Network Design for Image Restoration},
year = {2023},
booktitle = {Proceedings of the 40th International Conference on Machine Learning},
}

@inproceedings{focalnetcui2023focal,
  title={Focal Network for Image Restoration},
  author={Cui, Yuning and Ren, Wenqi and Cao, Xiaochun and Knoll, Alois},
  booktitle={Proceedings of the IEEE/CVF International Conference on Computer Vision},
  pages={13001--13011},
  year={2023}
}

@article{deanetchen2024dea,
  title={DEA-Net: Single image dehazing based on detail-enhanced convolution and content-guided attention},
  author={Chen, Zixuan and He, Zewei and Lu, Zhe-Ming},
  journal={IEEE Transactions on Image Processing},
  year={2024},
  publisher={IEEE}
}

@article{FDTANetgao2025frequency,
  title={Frequency domain task-adaptive network for restoring images with combined degradations},
  author={Gao, Hu and Ma, Bowen and Zhang, Ying and Yang, Jingfan and Yang, Jing and Dang, Depeng},
  journal={Pattern Recognition},
  volume={158},
  pages={111057},
  year={2025},
  publisher={Elsevier}
}

@inproceedings{PGH2Netisu2025prior,
  title={Prior-guided hierarchical harmonization network for efficient image dehazing},
  author={Su, Xiongfei and Li, Siyuan and Cui, Yuning and Cao, Miao and Zhang, Yulun and Chen, Zheng and Wu, Zongliang and Wang, Zedong and Zhang, Yuanlong and Yuan, Xin},
  booktitle={Proceedings of the AAAI Conference on Artificial Intelligence},
  volume={39},
  number={7},
  pages={7042--7050},
  year={2025}
}

@InProceedings{DefusionLuo_2025_CVPR,
    author    = {Luo, Wenyang and Qin, Haina and Chen, Zewen and Wang, Libin and Zheng, Dandan and Li, Yuming and Liu, Yufan and Li, Bing and Hu, Weiming},
    title     = {Visual-Instructed Degradation Diffusion for All-in-One Image Restoration},
    booktitle = {Proceedings of the IEEE/CVF Conference on Computer Vision and Pattern Recognition (CVPR)},
    month     = {June},
    year      = {2025},
    pages     = {12764-12777}
}

@InProceedings{VLUNetZeng_2025_CVPR,
    author    = {Zeng, Haijin and Wang, Xiangming and Chen, Yongyong and Su, Jingyong and Liu, Jie},
    title     = {Vision-Language Gradient Descent-driven All-in-One Deep Unfolding Networks},
    booktitle = {Proceedings of the IEEE/CVF Conference on Computer Vision and Pattern Recognition (CVPR)},
    month     = {June},
    year      = {2025},
    pages     = {7524-7533}
}

@article{kodak,
  title={Kodak lossless true color image suite},
  author={Franzen, Rich},
  journal={source: http://r0k. us/graphics/kodak},
  volume={4},
  number={2},
  pages={9},
  year={1999}
}
